# Comprehensive forecasting based analysis using stacked stateless and stateful Gated Recurrent Unit models

Swayamjit Saha*
*Department of Computer Science and Engineering*
*Swami Vivekananda Institute of Science & Technology, Kolkata, India*
swayamjit@ieee.org
swayamjit.saha97@gmail.com

Niladri Majumder*
*Department of Computer Science and Engineering*
*Swami Vivekananda Institute of Science & Technology, Kolkata, India*
niladrimajumder98@gmail.com

Devansh Sangani*
*Department of Computer Science and Engineering*
*Swami Vivekananda Institute of Science & Technology, Kolkata, India*
devanshsangani@gmail.com

** Authors with equal contribution*

## 1. Abstract

Photovoltaic power is a renewable source of energy which is highly used in industries. In economically struggling countries it can be a potential source of electric energy as other non-renewable resources are already exhausting. Now if installation of a photovoltaic cell in a region is done prior to research, it may not provide the desired energy output required for running that region. Hence forecasting is required which can elicit the output from a particular region considering its geometrical coordinates, solar parameter like GHI and weather parameters like temperature and wind speed etc. Our paper explores forecasting of solar irradiance on four such regions, out of which three is in West Bengal and one outside to depict with using stacked Gated Recurrent Unit (GRU) models. We have checked that stateful stacked gated recurrent unit model improves the prediction accuracy significantly.

## 2. Introduction

Due to the constant use of renewable resources for source of electricity, we're on the verge of losing our renewable resources count, thereby producing a scarcity of renewable resources. An approach to overcome this issue can be producing electrical power for our industry and domestic homes by utilizing Photovoltaic (PV) cells. As stated by Clack, C. T., in his paper [1] over the last decade the use of solar photovoltaics (PV) has expanded dramatically and that the deployment of solar PV has societal benefits, ranging from - no pollution from electric power production, very little water use, abundant resource, silent operation, long lifetime, and little maintenance. Solar irradiation is a promising source of energy due to large amount that the Earth receives daily, enough to supply on its own the needs of the entire planet. [3] Irradiation forecasting includes consideration of three solar parameters viz. Direct Normal Irradiance (DNI), Diffuse Horizontal Irradiance (DHI) and Global Horizontal Irradiance (GHI). Direct Normal Irradiance (DNI) is the aggregate of solar radiation received per unit area of a particular surface incident normally, and Diffuse Horizontal Irradiance (DHI) is the aggregate of radiation received per unit area of a particular surface that was scattered by different molecules of substances. Global Horizontal Irradiance (GHI) is the total amount of radiation received on a particular surface both normally and after getting scattered through the molecules, which is why we consider GHI for our experiment. Thus we tactfully evade from the iteration of using DNI and DHI all over again, as GHI considers both DNI and DHI for the individual calculation. GHI forecasting gives statistically significant results based on the regions we have covered. M. C. Sorkun et al. in their paper [2] uses machine learning models to forecast time series solar irradiation data.

A Gated Recurrent Unit (GRU) introduced by Cho, K. et al. in their paper [4], is a variation of RNN architecture which employs gates to control the flow of information between the various cells of an unit. GRU is a new model as compared to LSTM architecture introduced by Hochreiter, S. et al in their paper. [9] The ability of GRU to adhere to long-term dependence or memory from the internal direction of the GRU cell to produce a hidden state. While LSTMs have two different states that have passed between cells - cell status and hidden state, with long and short memory, respectively - only GRUs have one hidden state transferred between time intervals. This hidden state is capable of capturing long-term and short-term reliability due to hidden processing methods and accounting and input data. The GRU cell contains only two gates: the Update gate and the Reset gate. Like gates on LSTMs, these GRU gates are trained to filter out any inaccurate information while maintaining usability. These gates are basically vans containing 0 to 1 values that will be expanded with input data and / or hidden country. A value of 0 on the gate vases indicates that the corresponding data in the installation or hidden region is insignificant and, therefore, will return as zero. On the other hand, the number 1 in the image at the gate means that the corresponding data is important and will be used. In this paper the terms gate and vector are synonymous and will be used interchangeably. The structure of a GRU unit is shown in Figure 1.

## 3. Background:

Recurrent Neural Networks (RNN) suffer from vanishing gradient and exploding problems which hampers learning of long data sequences. When using tanh or relu functions, RNN often cannot process the long sequences, making training process a difficult job. Gated Recurrent Unit (GRU) acts as an improved version of RNN by solving the vanishing gradient problem. As can be predicted from the name, GRU uses two technical gates viz. Update Gate and Reset Gate which controls the information to be passed to the output. GRUs thus experience faster execution time as compared to LSTM. As compared to LSTM recurrent neural networks, GRUs have a different architecture with inclusion of two new gate vectors viz. Reset gate managing the amount of new memory to be added and Update gate for determining which information to retain from the previous states.GRU is attributed with fewer number of weights which results in faster training time as compared to the LSTM RNN model.

Time series analysis is the gathering of information at specific intervals over a definite period of time with the sole purpose of identifying cycles, trends or seasonal variances to assist in forecasting of a future event. Our paper explores the different exposures of solar irradiance in several parts of the state of West Bengal, India having tropical type of climate and we compare the results with state Orissa, India having a tropical Savannah type of climate influenced by sea breeze.

Our paper deals with hourly basis of time series data of eight regions viz. Orissa, Malda, Darjeeling, Burdwan, Murshidabad, Baruipur, Purulia and Raiganj. On addition to that, three different months viz. January, July and October data are taken which are supposed to have different exposures to parameters like solar irradiance, temperature, pressure, relative humidity, wind speed and solar zenith angle. The previous papers have not considered so many parameters for experimenting with Gated Recurrent Unit model. Our paper comprehensively forecasts data with the help of two unique models of stacked gated recurrent unit viz. Stateful Stacked GRU and Stateless Stacked GRU. Our analysis shows that Stateful GRU performs better in forecasting irradiance than Stateless GRU in majority of the regions.

## 4. Approach

### 4.1 Flow chart

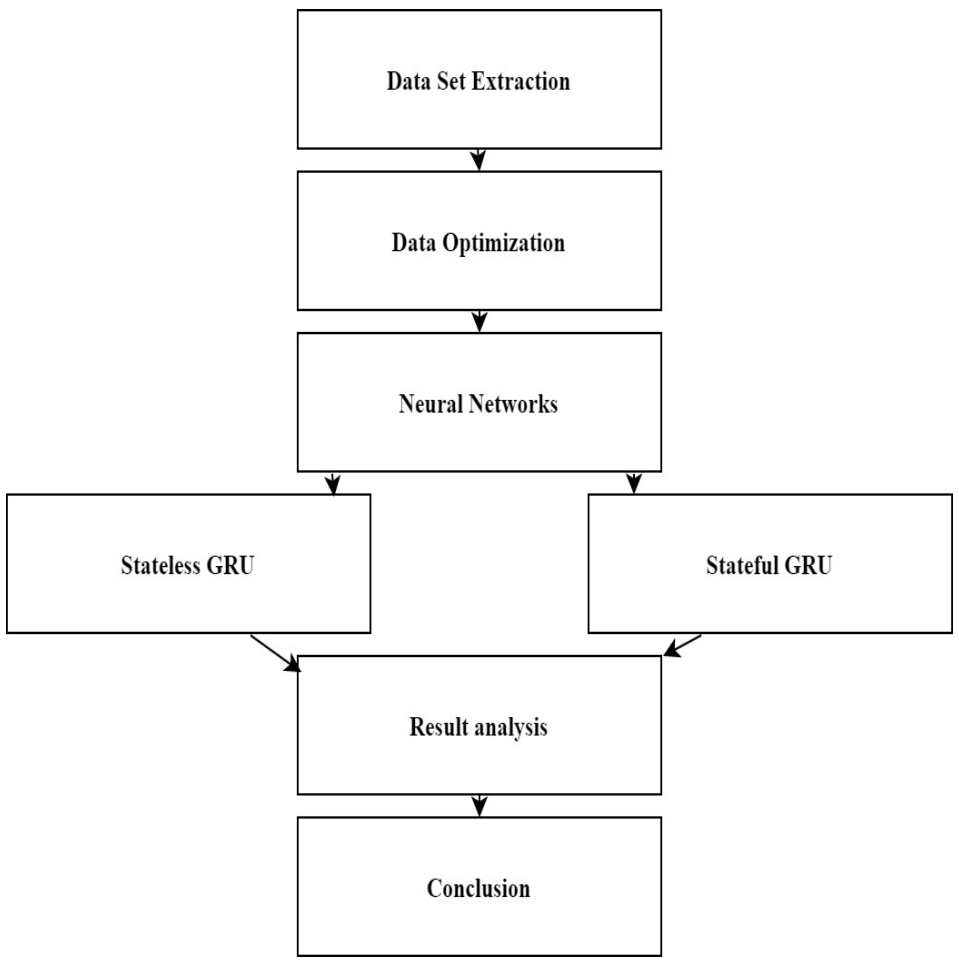

*A. Flow- chart description*

Data Set Extraction: The dataset is extracted from the SURFace RADiation budget (SURFRAD) network (https://www.esrl.noaa.gov/gmd/grad/surfrad/) which is publicly available.

Data Optimization: Raw data consists of iterative parameters viz. DNI, DHI which were ignored for our experiment. We considered forecasted parameters viz. Hour, GHI and forecasting parameters viz. temperature, pressure, relative humidity, wind speed and solar zenith angle in order to get accurate predictions.

Neural Networks: The optimized data is trained with neural networks. The selected recurrent neural networks are: Stateless Staked Gated Recurrent Unit and Stateful Staked Gated Recurrent Unit.

Result analysis: The results from the two recurrent neural networks are collected and then compared. This is done to show which model performs better for forecasting solar irradiance.

Conclusion: This stage determines which model performs the best among the two models chosen. Finally, we present the algorithm for the best-proven model for forecasting solar irradiance.

### 4.2 Dataset

We believe data accumulation is an important procedure in the conducted experiment. To produce accurate solar irradiance estimates the use of excellent quality solar measurements is fundamental. The United States has many such high quality measurement networks. Two of them are used in the present paper: the SURFace RADiation budget (SURFRAD) network [http://www.esrl.noaa.gov/gmd/grad/surfrad/] and the Integrated Surface Irradiance Study (ISIS) Network [http://www.esrl.noaa.gov/gmd/grad/isis/]. [6-8] All of the data are publicly available.

The dataset consists of 3720 number of sample records and each of which can be framed as collection of attributes that include several criterions for forecasting power generation in the selected regions. The dataset can be formulated as collection of attributes that include several criterions for forecasting such as Pressure, Relative humidity, solar zenith angle (angle between the zenith and the centre of the Sun's disc), Temperature (°C) and Wind speed (knot). The other forecasted parameters are GHI (Global Horizontal Irradiance) DHI (Diffuse Horizontal Irradiance) and DNI (Direct Normal Irradiance) which are used to evaluate the power generated.

### 4.3 Network architecture

Gated Recurrent Unit (GRU) is a facile model that generally has a low capability of feature extraction. Stacked GRU on the contrary is robust in structure composing several GRUs in a single unit shown in Figure 2.

Gated Recurrent Unit (GRU) is a modification to the RNN hidden layer that makes it much better capturing long range connections and Vanishing Gradient Problem.

$b^{<t>} = f(X_b[b^{<t-1>}, x^t] + c_b)$....................................... (1)

$[b^{<t-1>}, x^t]$– previous activation time set

f is Activation function (tanh).

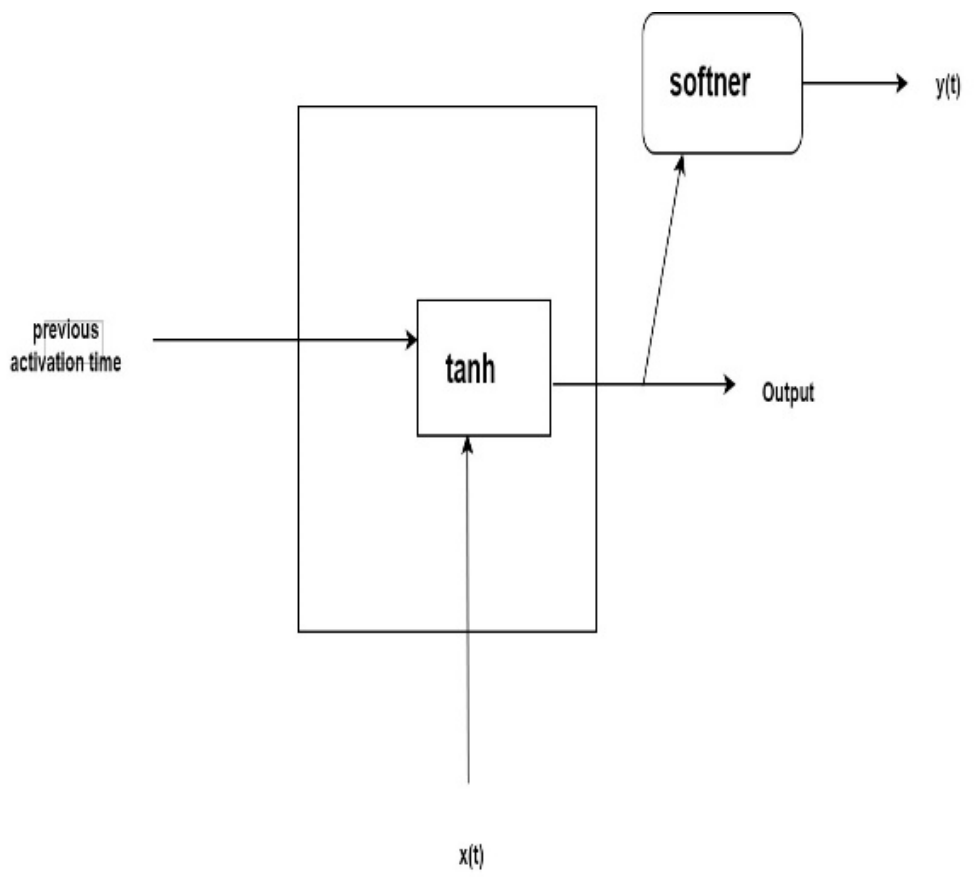

**Figure 1 A Gated Recurrent Unit model**

The output $b^{<t>}$can be calculated by including the inputs of previous activation time step $b^{<t-1>}$and $x^{t}$ , $X_{b}$times.

Now we consider a new variable for memory cell $c_m$.

Let us consider a sample statement: The child, which already ate… was full. [5]

We know an equation

$c^{<t>}= b^{<t>}$…………………………………. (2).

$c'^{<t>}$= tanh $(X_c[c^{<t-1>}, x^{<t>}] + b_a$ [ where, $b_a$ =bias, c' is candidate for replacing c]

The idea of GRU is to introduce a Gate hence we use an Update Gate.

$g_{up} = \sigma\ (X_c[c^{<t-1>}, x^{<t>}] + b_a\ )$…………………………….. (3)

σ = Sigmoid function

$g_{up}$ = update gate variable

We then try to decide whether c is singular or plural. The GRU unit will try to memorise the value of c from previous unit.

The job of $g_{up}$ is to determine whether we should update the value or not.

$c^{<t>}= g_{up} * c'^{<t>} + (1 - g_{up})* c^{<t-1>}$………………… (4)

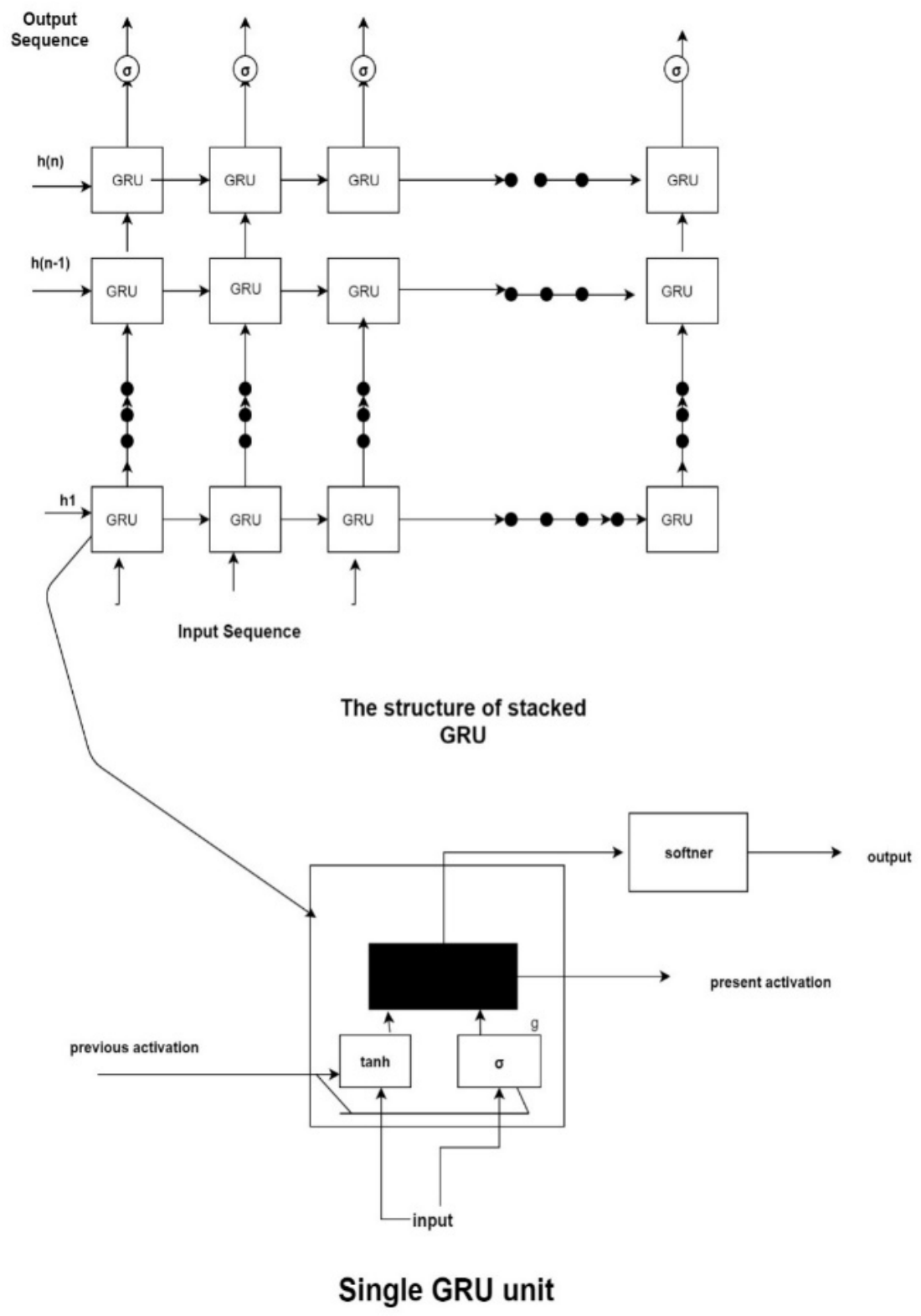

**Figure 2 Stacked Gated Recurrent Unit model**

Here $g_{up}$ equals 1 as “The child” here is singular.

Now for all remaining instances $g_{up}$ will not update the values of $c^{<t>}$.

So even if value of $g_{up}$ gets almost equal to 0, it still holds the value of previous c i.e. $c^{<t-1>}$. This happens because we have the part $(1 - g_{up})$as it will get almost equal to 1(g almost 0). So new value of c becomes $c^{<t-1>}$. Hence the vanishing gradient problem is solved. RNN is hence used for capturing long range connections.

## 5. Results

Root Mean Square Error (RMSE) is a standard loss function to evaluate model test performance for regression. We compare Stacked Stateful Gated Recurrent Unit with Stacked Stateless Gated Recurrent Unit models for forecasting future solar irradiation for a 24-hour time stamp basis. Four regions are considered, out of which three lies typically in West Bengal, India and one outside West Bengal. This is done carefully to understand how efficient the models are performing. Minimum RMSE is highlighted in bold. See Table 1.

In neural networks, loss is used to define the prediction error. The loss and value loss values for Stateless Stacked GRU and Stateful Stacked GRU are given in Table 2 and Table 3.

The average performance indicates that for four regions, Stacked Stateful GRU minimizes the RMSE than the other models. The same can be seen in Table 4, where stacked stateful GRU is performing better for all the three months January, July and October. Table 4 shows their detailed performance for all the four selected regions.

The loss curves for Stateless Stacked GRU gives an idea about how well the model performs during testing, after it is trained with data.

Table **5** gives their detailed performance for all the four regions. Table 6 gives RMSE details for Stateless Stacked GRU.

## 6. Conclusion

We introduced stacked gated recurrent unit models for forecasting solar irradiance data and compared it with the respective stateless and stateful models. It was found that stateful stacked gated recurrent unit model is suitable to minimize the RMSE. The experimental results also show that stateless stacked gated recurrent unit is not suitable for reducing RMSE loss. We additionally provide multi-location forecasting to analyze how the models vary with respect to performance.

**Table 1 Results of monthly accumulated RMSE data for the selected regions**

| City | Latitude | Longitude | Month | Train RMSE | | Test RMSE | |
|---|---|---|---|---|---|---|---|
| | | | | Stateless GRU | Stateful GRU | Stateless GRU | Stateful GRU |
| Orissa | 22.85 | 80.75 | January | 55.37 | **55** | **32.76** | 35.48 |
| | | | July | 106.88 | **105.08** | 103.62 | **103.24** |
| | | | October | 83.05 | **80.15** | 92.45 | **91.71** |
| Darjeeling | 27.05 | 88.25 | January | **68.84** | 69.74 | **72.65** | 76.14 |
| | | | July | 91.95 | **89.76** | **115.51** | 118.50 |
| | | | October | 85.72 | **83.60** | **84.92** | 89.72 |
| Baruipur | 22.35 | 88.45 | January | **51.45** | 57.85 | **43.18** | 55.46 |
| | | | July | 110.66 | **109.92** | **107.19** | 109.13 |
| | | | October | 80.33 | **78.48** | 89.21 | **86.81** |
| Burdwan | 23.45 | 88.35 | January | 54.86 | **53.96** | 60.17 | **58.85** |
| | | | July | 101.03 | **99.90** | 102.35 | **101.69** |
| | | | October | 80.33 | **68.22** | **89.21** | 90.85 |

Note: Decreased value of Train RMSE is an useful sanity check. Test RMSE decreased values are used to predict how well the models are fitted to the training data.

**Table 2 Results of monthly accumulated loss at each location**

| | Loss | | | | | |
|---|---|---|---|---|---|---|
| | Stateless GRU | | | Stateful GRU | | |
| **City** | **January** | **July** | **October** | **January** | **July** | **October** |
| Orissa | 0.0047 | 0.0124 | 0.0084 | 0.0048 | 0.0122 | 0.0080 |
| Darjeeling | 0.0080 | 0.0086 | 0.0093 | 0.0085 | 0.0086 | 0.0087 |
| Baruipur | 0.0044 | 0.0133 | 0.0087 | 0.0043 | 0.0133 | 0.0088 |
| Burdwan | 0.0052 | 0.0112 | 0.0060 | 0.0053 | 0.0111 | 0.0062 |

Note: Decreased values of loss illustrates good prediction of the arbitrary testing data.

**Table 3 Results of monthly accumulated value loss at each location**

| | Value Loss | | | | | |
|---|---|---|---|---|---|---|
| | Stateless GRU | | | Stateful GRU | | |
| **City** | **January** | **July** | **October** | **January** | **July** | **October** |
| Orissa | 0.0016 | 0.0117 | 0.0104 | 0.0022 | 0.0118 | 0.0111 |
| Darjeeling | 0.0088 | 0.0136 | 0.0090 | 0.0096 | 0.0143 | 0.0101 |
| Baruipur | 0.0131 | 0.0125 | 0.0107 | 0.0051 | 0.0130 | 0.0102 |
| Burdwan | 0.0063 | 0.0115 | 0.0100 | 0.0060 | 0.0113 | 0.0113 |

Note: Decreased value losses illustrates good prediction of the arbitrary testing data.

**Table 4 Stacked Stateful GRU Hour vs GHI plots**

| **Cities** | **Selected Months** | | |
|---|---|---|---|
| | January | July | October |
| Orissa | 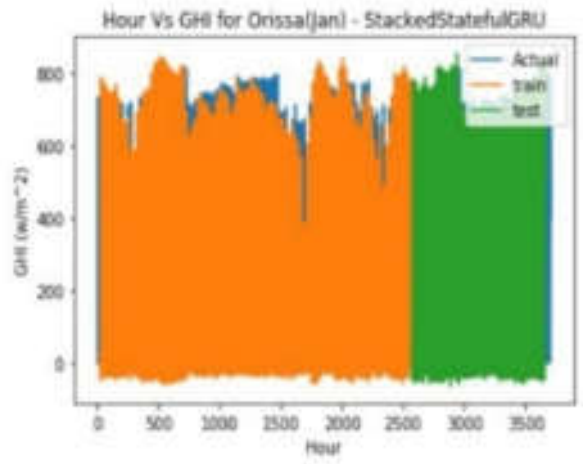 | 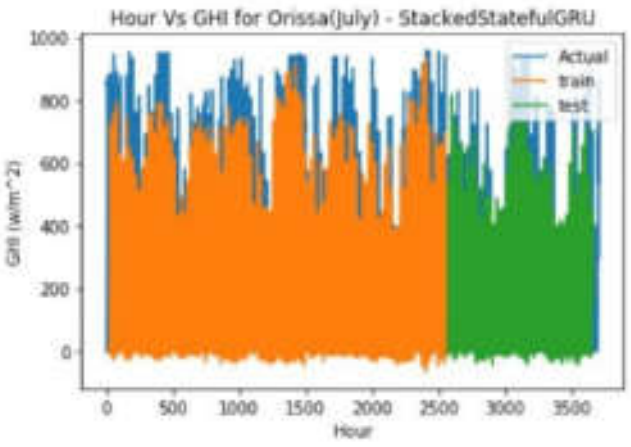 | 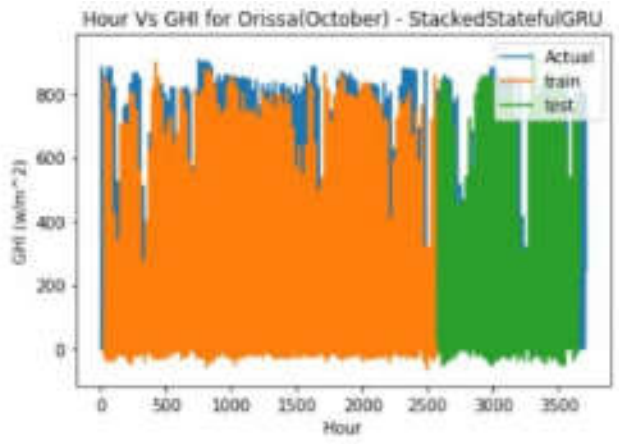 |
| Darjeelin-g | 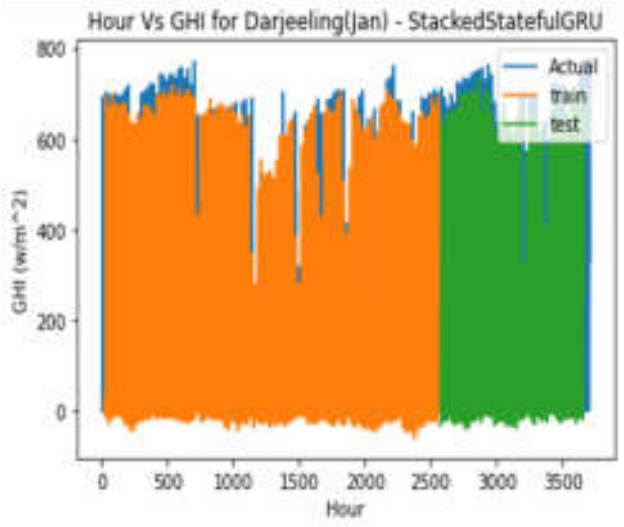 | 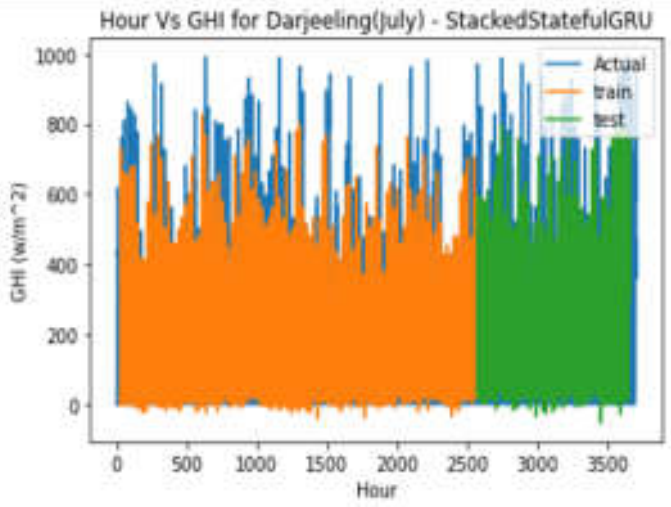 | 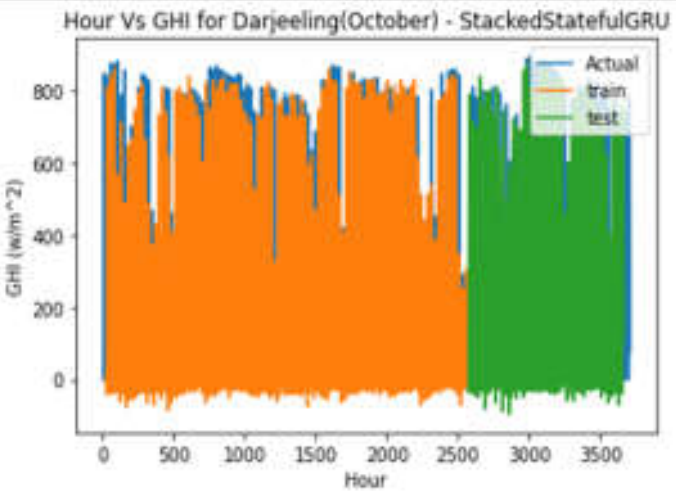 |
| Baruipur | 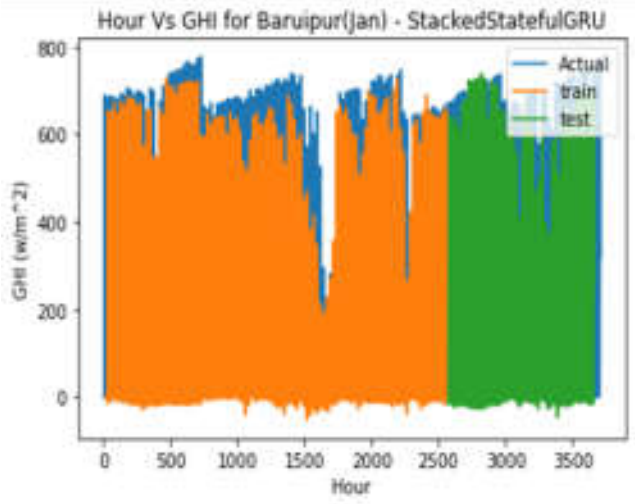 | 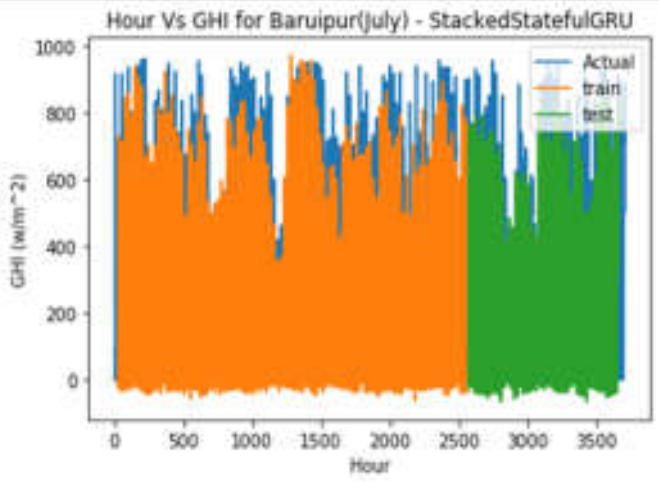 | 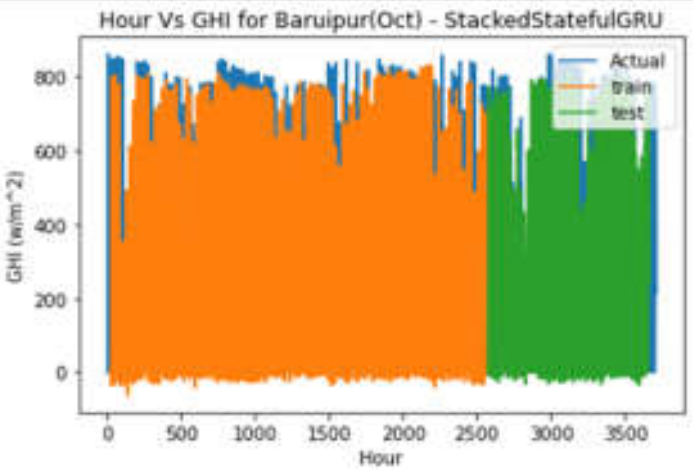 |
| Burdwan | 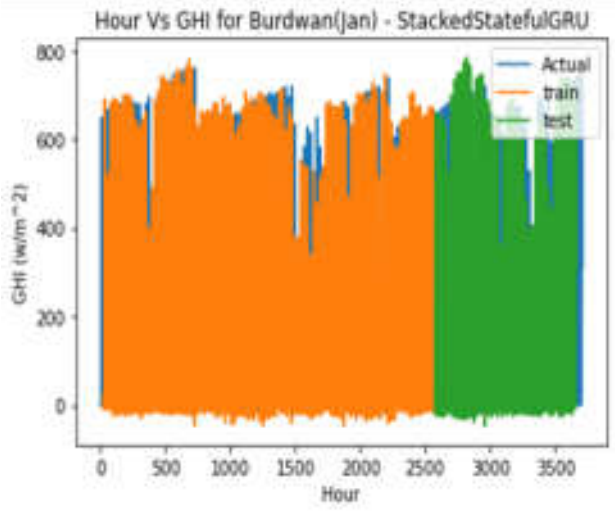 | 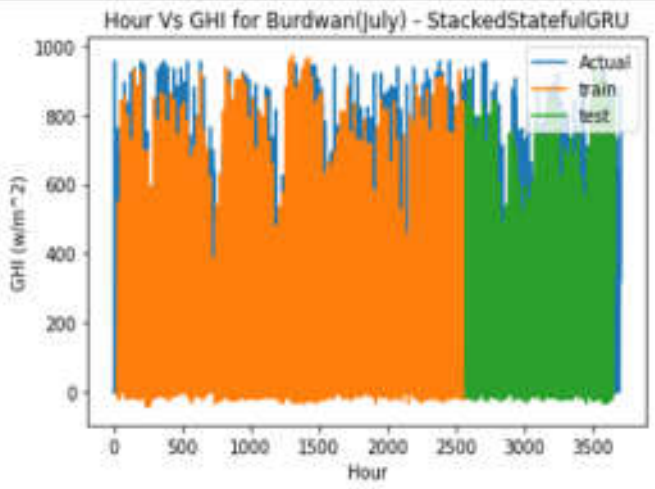 | 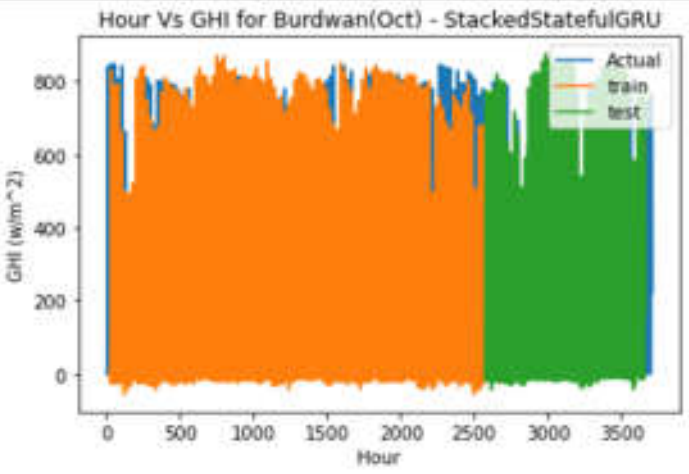 |

**Table 5 Model loss curves for the selected regions**

<table>
<tr><th rowspan="2">Cities</th><th colspan="3">Selected Months</th></tr>
<tr><td>January</td><td>July</td><td>October</td></tr>
<tr><td>Orissa</td><td>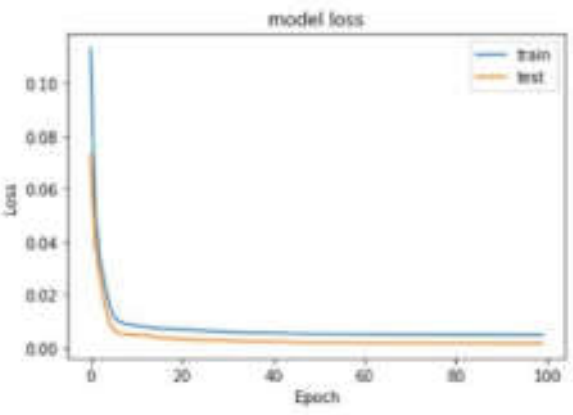
</td><td>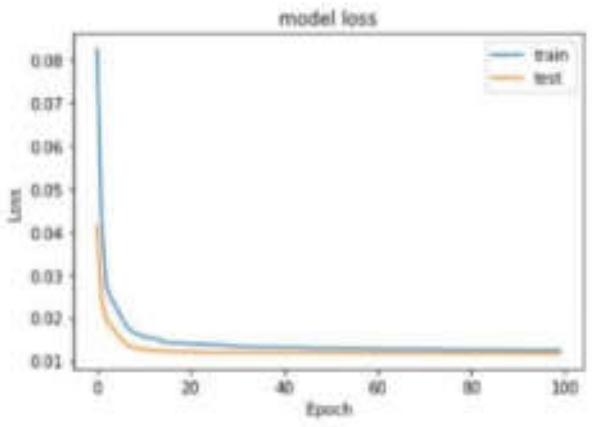
</td><td>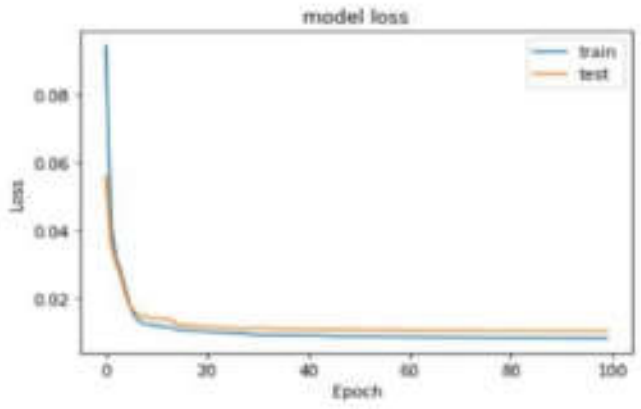
</td></tr>
<tr><td>Darjeeli-<br>ng</td><td>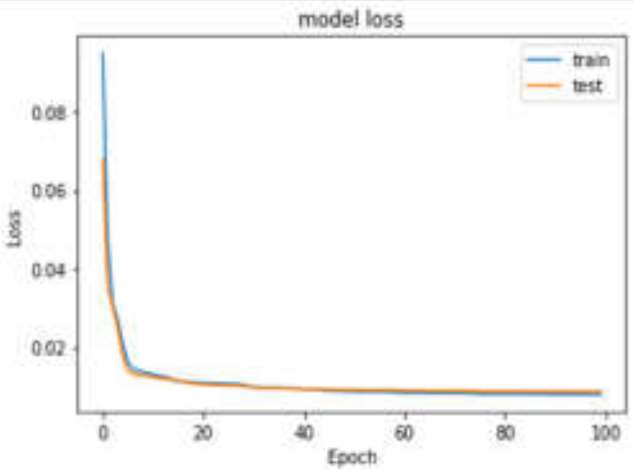
</td><td>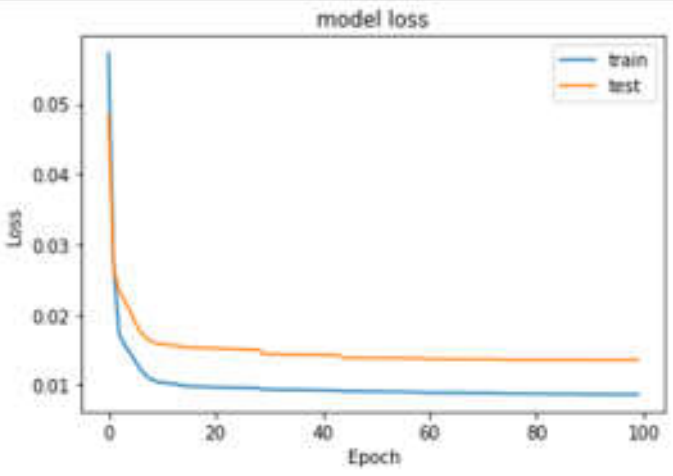
</td><td>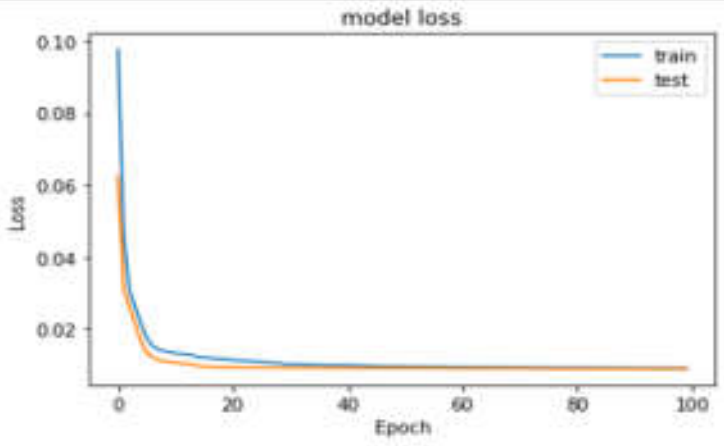
</td></tr>
<tr><td>Baruipu-<br>r</td><td>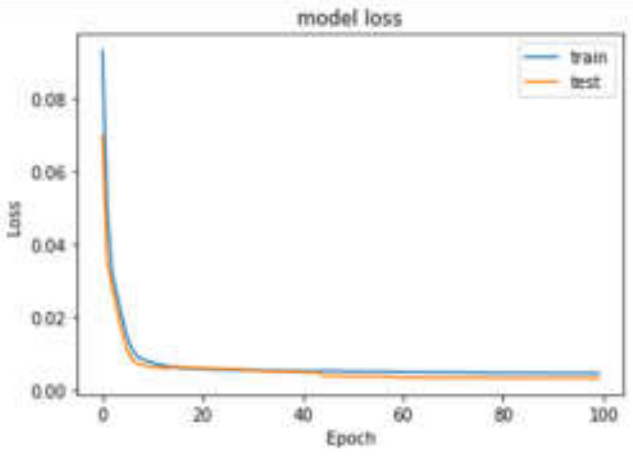
</td><td>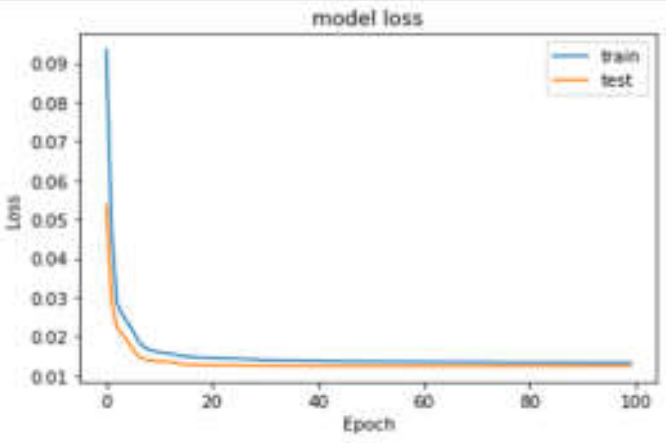
</td><td>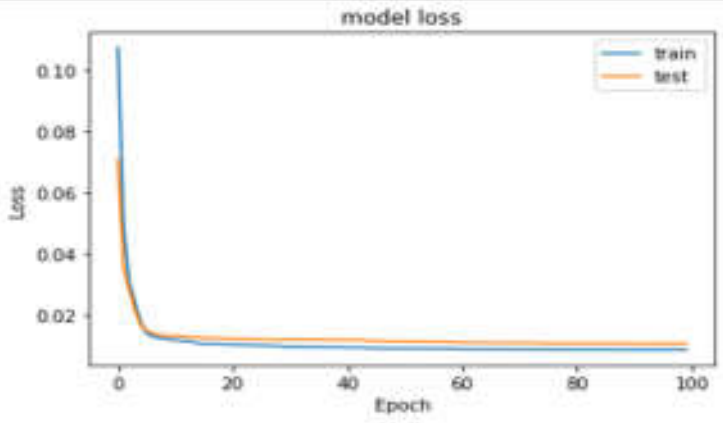
</td></tr>
<tr><td>Burdwa-<br>n</td><td>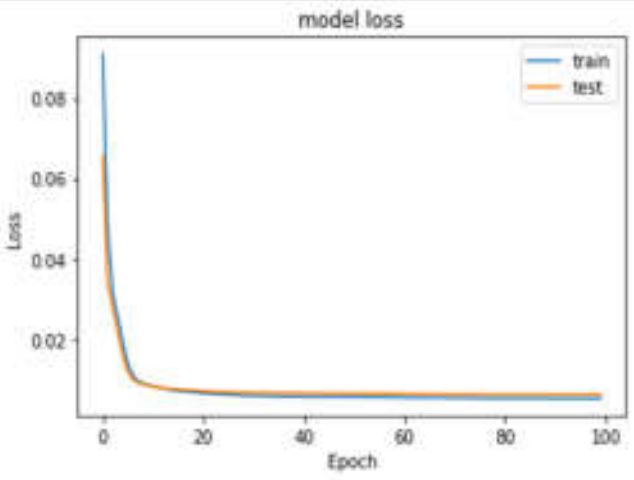
</td><td>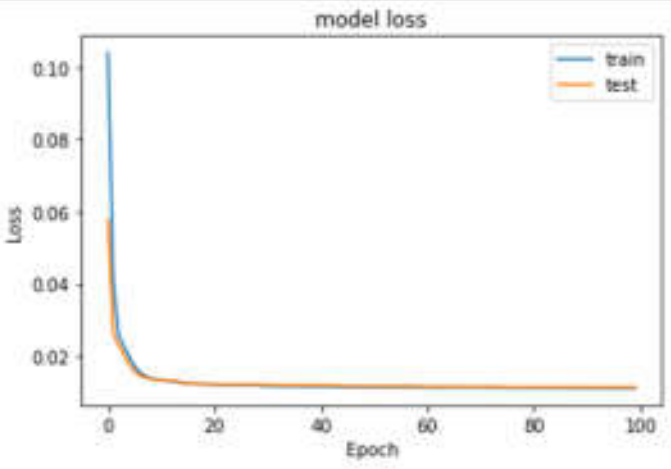
</td><td>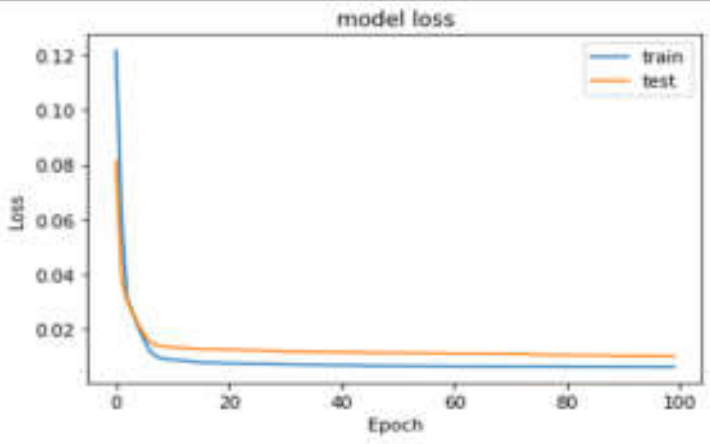
</td></tr>
</table>

**Table 6 Stacked Stateless GRU Hour vs GHI plots for the selected regions**

| Cities | Selected Months | | |
|---|---|---|---|
| | January | July | October |
| Orissa | 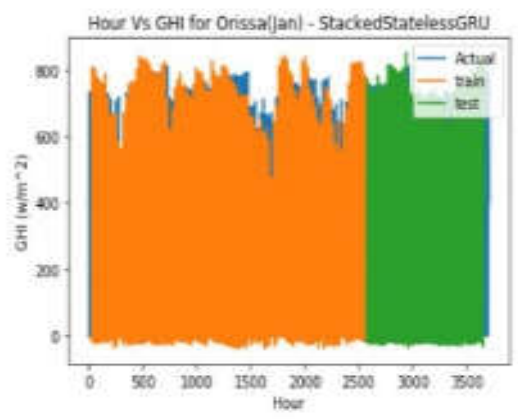 | 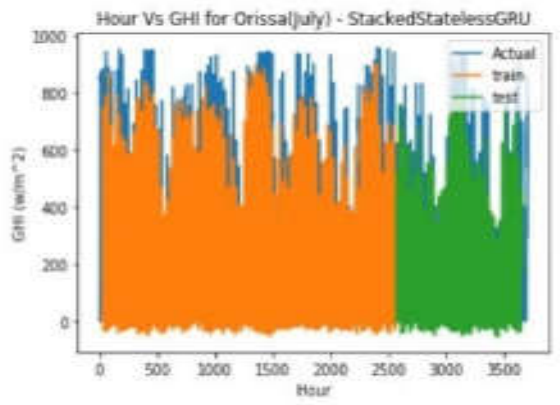 | 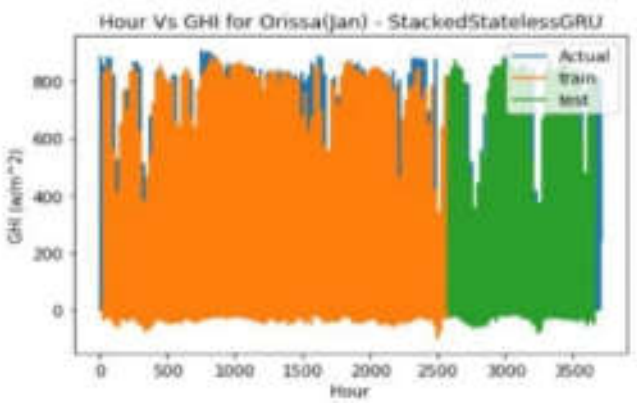 |
| Darjeeling | 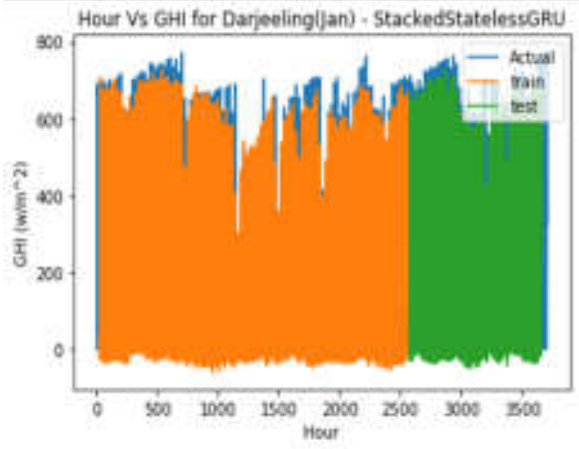 | 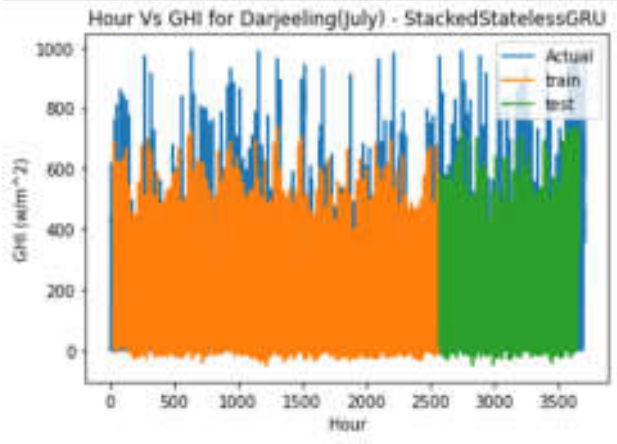 | 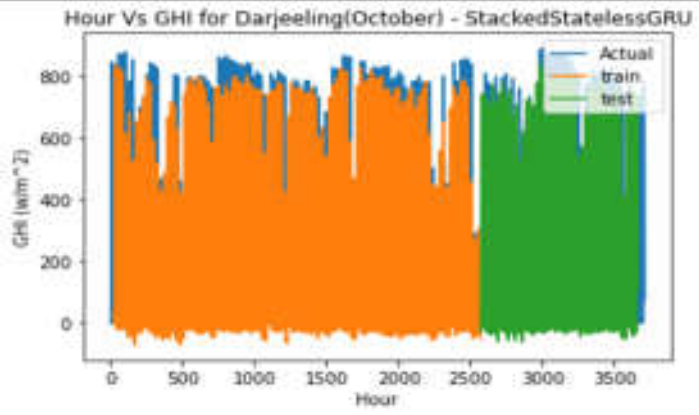 |
| Baruipur | 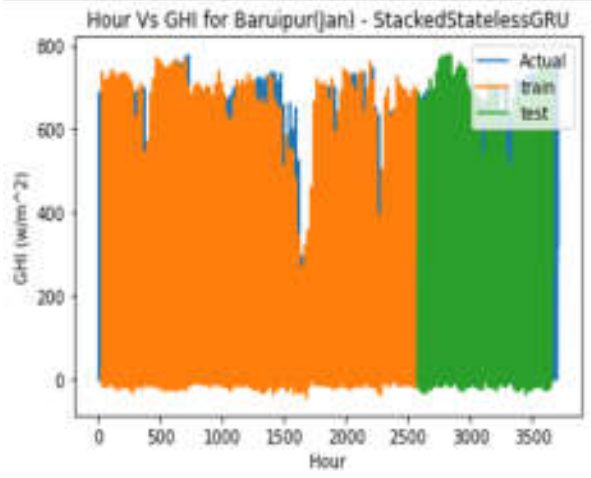 | 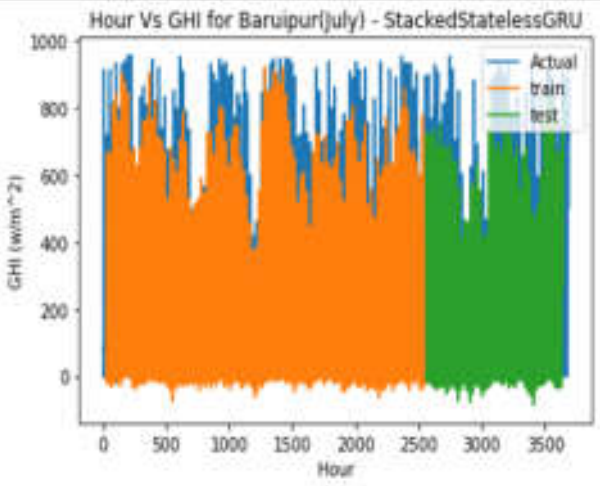 | 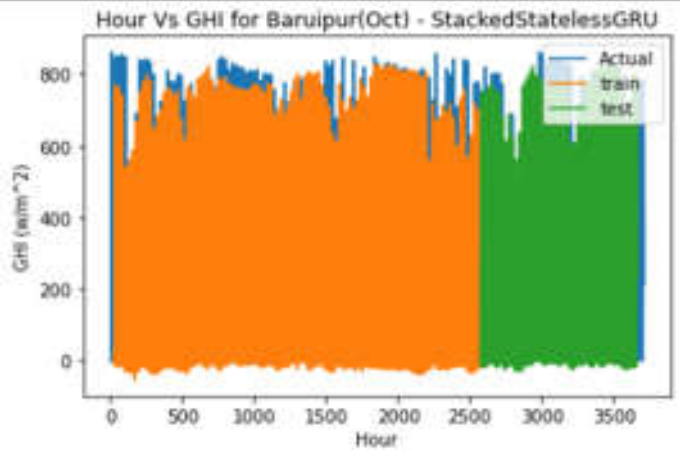 |
| Burdwan | 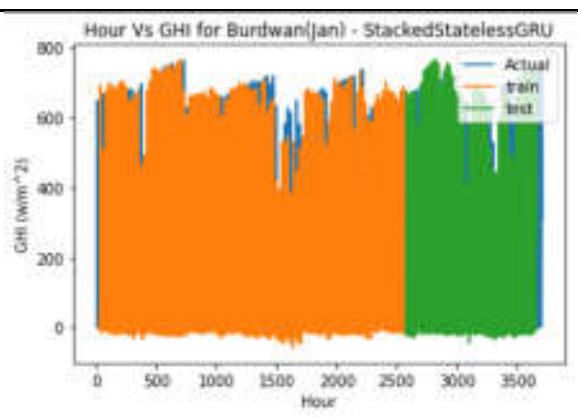 | 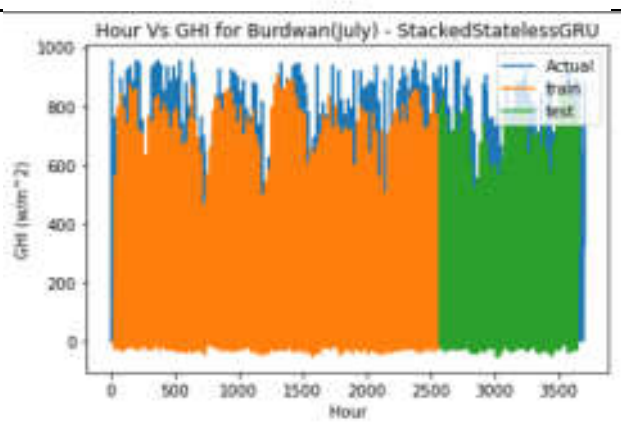 | 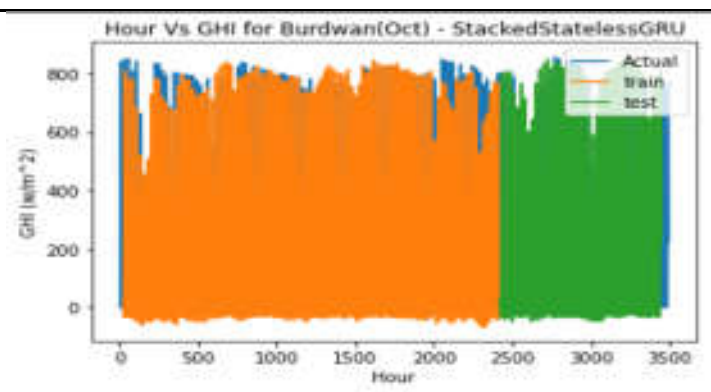 |